\documentclass[letterpaper, 10 pt, conference]{ieeeconf} 
\IEEEoverridecommandlockouts
\usepackage{graphics} 
\usepackage{epsfig} 
\usepackage{amsmath} 
\usepackage{amssymb} 
\usepackage{color}
\usepackage{bm}
\usepackage{bbm}
\usepackage{subcaption}
\usepackage{multirow}
\usepackage{array}
\usepackage{booktabs}
\usepackage{algorithm,algcompatible}
\usepackage{mathtools}
\usepackage{cite}
\usepackage{hyperref}
\usepackage{cuted}
\definecolor{pointcolor}{RGB}{0,114,189}
\definecolor{linecolor}{RGB}{217,83,25}
\definecolor{planecolor}{RGB}{237,177,32}
\definecolor{spherecolor}{RGB}{119,172,48}
\definecolor{ellipsoidcolor}{RGB}{126,47,142}
\definecolor{cylindercolor}{RGB}{77,190,238}
\definecolor{conecolor}{RGB}{162,20,47}

\graphicspath{{figures/}}

\title{\huge \bf QuadricsNet: Learning Concise Representation for Geometric Primitives in Point Clouds}

\author{Ji Wu$^*$\quad  Huai Yu$^*$\quad  Wen Yang\quad  Gui-Song Xia
\thanks{$^*$ These authors contribute equally to this work.}
\thanks{Ji Wu and Gui-Song Xia are with the School of Computer Science, Huai Yu and Wen Yang are with the School of Electronic Information. All authors are with Wuhan University, Wuhan, China 430072. {\tt\small\{ji.wu,yuhuai, yangwen,guisong.xia\}@whu.edu.cn}}
}

\begin{document}

\maketitle
\begin{abstract}
This paper presents a novel framework to learn a concise geometric primitive representation for 3D point clouds. Different from representing each type of primitive individually, we focus on the challenging problem of how to achieve a concise and uniform representation robustly. We employ \textit{quadrics} to represent diverse primitives with only 10 parameters and propose the first end-to-end learning-based framework, namely \textit{QuadricsNet}, to parse \textit{quadrics} in point clouds. The relationships between \textit{quadrics} mathematical formulation and geometric attributes, including the \textit{type}, \textit{scale} and \textit{pose}, are insightfully integrated for effective supervision of \textit{QuaidricsNet}. Besides, a novel pattern-comprehensive dataset with \textit{quadrics} segments and objects is collected for training and evaluation. Experiments demonstrate the effectiveness of our concise representation and the robustness of \textit{QuadricsNet}. Our code is available at \url{https://github.com/MichaelWu99-lab/QuadricsNet}.

%

\end{abstract}
\section{Introduction}

Geometric primitive representation and parsing are fundamental problems for compact 3D object representation and scene modeling, providing crucial features for structured mapping \cite{nan2017polyfit,monszpart2015rapter,wang2020robust}, CAD reverse engineering \cite{benko2002constrained,uy2022point2cyl}, and SLAM optimization \cite{kaess2015simultaneous,zhen2022unified,taguchi2013point}. Unlike non-vectorized, sparse and large point clouds, geometric primitives can model the geometric information in a compact manner with vectorized representation. Several geometric primitives have proven to be successful in various fields during the last decades, such as planes \cite{sommer2020planes}, ellipsoids \cite{nicholson2018quadricslam}, B-splines \cite{DBLP:conf/eccv/SharmaLMKCM20} and cuboids \cite{taguchi2013point}. However, unified representation and modeling of these primitives remain a challenging problem. Moreover, although deep learning has achieved great success in 2D and 3D object detection and classification, end-to-end unified mathematical modeling and geometric perception are pressing issues that need to be explored when parsing geometric primitives.

\begin{figure}[t]
  \centering
   \includegraphics[width=1\linewidth]{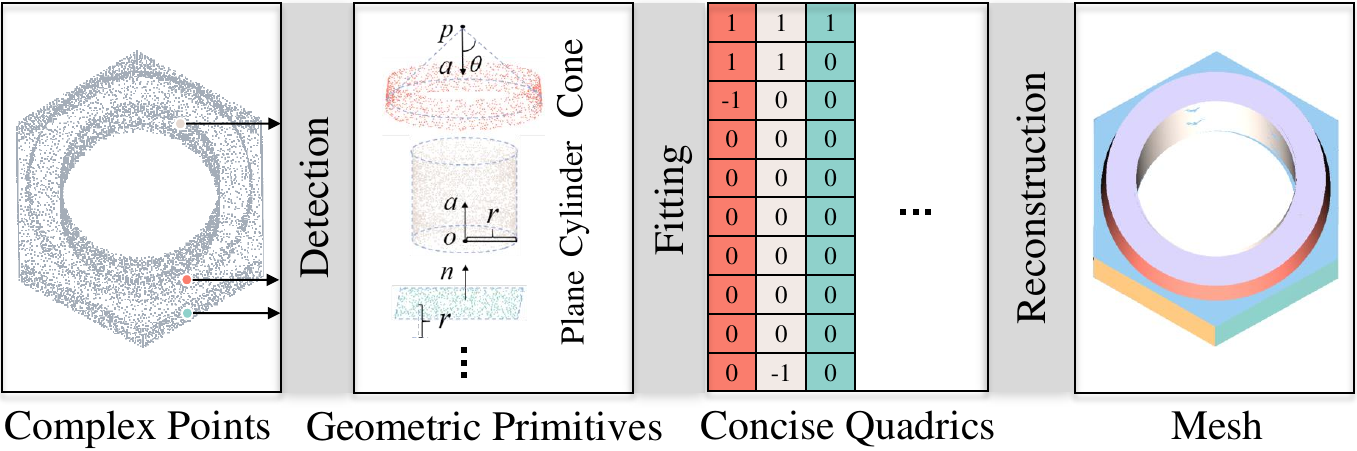}
   \caption{For point clouds with various geometric primitives, representing them individually poses a challenge for the algorithm design and downstream tasks. We propose \textit{QuadricsNet} to learn a concise representation with 10 parameters for diverse geometric primitives, thus yielding robust primitive parsing and structure mapping results.}
   \label{fig:1-1}
\end{figure}

Real-world objects and scenes are generally composed of multiple geometric elements. For example, a cup typically consists of a cylinder and two planes, while streets usually include planes, cylinders, and cones. As shown in Fig. \ref{fig:1-1}, most methods employ one or several geometric primitives for simplicity \cite{DBLP:conf/cvpr/LiSDYG19,DBLP:conf/eccv/SharmaLMKCM20,yan2021hpnet,le2021cpfn}. Eiffient-RANSAC \cite{schnabel2007efficient} integrated in CGAL is a standard analytical algorithm. However, the configurations must be carefully tuned to accommodate different types of primitives. Learning-based techniques can obtain more robust configurations from massive training data \cite{DBLP:conf/cvpr/LiSDYG19}. Nevertheless, each primitive is still represented individually, which requires the design of different learning models for different primitives. Therefore, if we can unify the representation of different primitives, it will significantly simplify the design complexity and parameter settings of parsing methods, thereby improving the robustness and generalization of primitive parsing. 

For the unified representation and modeling, we find that \textit{quadrics} in 3D space can concisely represent 17 types of geometric primitives with only 10 parameters, which has been introduced into 3D vision for abstract mapping \cite{birdal2019generic} and SLAM \cite{zhen2022unified}. However, existing methods often use non-learning geometric clustering and fitting strategies, which require high-fidelity point clouds and careful parameter configurations. Point cloud noises and parameter errors will lead to fragmented segmentations and inaccurate primitive fitting. Deep learning models have stronger adaptability than traditional methods in solving data noise and parameter fitting. With a large amount of training data, an intuition is that using deep models to parse these primitives may have better performance than traditional methods. However, the learning of the quadrics representation for parsing geometric primitives remains unexplored. 
 
In this work, we propose a robust learning-based quadrics detection and fitting framework for the parsing of geometric primitives in point clouds, namely \textit{QuadircsNet}. It consists of a quadrics detection module and a quadrics fitting module. The detection module segments and classifies point clouds into several quadrics primitives, while the fitting module generates quadrics parameters from the detected primitives and vectorizes the primitive models. Our key insight is utilizing the characteristics obtained from quadrics mathematical formulation, such as \textit{type}, \textit{scale}, and \textit{pose}, to supervise the quadrics detection and fitting modules. The key contributions can be summarized as:

\begin{itemize}
\item[$\bullet$] We propose \textit{QuadricsNet}, the first end-to-end quadrics detection and fitting framework for learning the concise representation of geometric primitives in point clouds. 
\item[$\bullet$] We leverage the geometric attributes from quadrics mathematical decomposition, such as the \textit{type}, \textit{scale}, and \textit{pose}, to supervise the detection and fitting modules.
\item[$\bullet$] We build \textit{QuadricsDataset}, which consists of quadrics segments and objects. Experiments demonstrate the effectiveness of our quadrics representation and robustness of \textit{QuadricsNet}.
\end{itemize}

\section{Related Work}
In this section, we review the representations of geometric primitives and the detection-fitting techniques.
\subsection{Geometric Primitive Representation}
The representation of geometric primitives has been widely investigated in computer vision and robotics \cite{9293398}. Planes receive the earliest attention as the simplest primitive, which are represented as a normal and an offset in most methods \cite{borrmann20113d,fang2018planar,sommer2020planes}. Further, curved surfaces such as cylinders, spheres, and cones are represented by the parameters like apex, axis, or radius, depending on their geometric structures \cite{DBLP:conf/cvpr/LiSDYG19,schnabel2007efficient,li2011globfit}. There are also more particular surfaces, \textit{e.g.}, B-spline \cite{DBLP:conf/eccv/SharmaLMKCM20} and extrusion cylinders \cite{uy2022point2cyl}, which are also modeled according to their properties. The above individual representations lack uniformity, resulting in the necessity of multiple designs for detecting diverse primitives, as well as inconveniences for downstream tasks.

A more general representation could address the above deficiencies. For example, cuboids \cite{taguchi2013point,8708251} are used to represent objects as bounding boxes. Ellipsoids \cite{nicholson2018quadricslam} and superquadircs \cite{paschalidou2019superquadrics} have a better expression in terms of scale, orientation, and position. But they are limited by their basic models, resulting in only being able to express a coarse structure. Surfaces based on control points \cite{farin2002curves,DBLP:conf/eccv/SharmaLMKCM20} can express arbitrary shapes. However, the parameters of the control points are redundant and have insufficient geometric attributes. In contrast, \textit{quadrics} can concisely cover the 17 most common geometric primitives, and can also figure out their attributes, such as \textit{type}, \textit{scale}, and \textit{pose} \cite{birdal2019generic,zhen2022unified}. However, the relationships between quadrics mathematical formulation and geometric attributes are not fully considered, and non-learning methods suffer from weak robustness and generalization.

\subsection{Detection and Fitting of Primitives}
Detection and fitting of geometric primitives in point clouds is a time-honored problem \cite{kaiser2019survey}. Hough transforms based on voting in parameter space \cite{borrmann20113d,drost2015local,sommer2020primitect}, region growing based on similarity matching \cite{oesau2016planar}, and iterative heuristic RANSAC \cite{fischler1981random,schnabel2007efficient,li2011globfit} have achieved great success and continuous extension. Particularly, Efficient-RANSAC \cite{schnabel2007efficient} implemented in CGAL is considered the standard. However, the non-learning methods suffer from laborious tuning when confronted with different primitives. In addition, noise and challenging cases like fragmentation and occlusions are also intractable. 

Learning-based methods are highly anticipated in these issues. SPFN \cite{DBLP:conf/cvpr/LiSDYG19} first detects four primitives and then fits the corresponding parameters with separate differentiable estimators. ParseNet \cite{DBLP:conf/eccv/SharmaLMKCM20} and Point2Cyl \cite{uy2022point2cyl} extend the primitive types to B-splines and extrusion cylinders. HPNet \cite{yan2021hpnet} proposes hybrid features to combine multiple primitive detection cues, thereby improving detection performances. CPFN \cite{le2021cpfn} assembles the detection results of global and local networks by an adaptive patch sampling network to improve the detection of fine-scale primitives. However, these networks are designed with individual estimators for specific primitives. There are only a few networks for general primitives \cite{tulsiani2017learning,zou20173d,paschalidou2019superquadrics}, but they are limited by the expressiveness of the representations used and hence can only represent structures at a coarse level.

Different from prior works, we are the first to adopt a learning approach to represent point clouds with \textit{quadrics} concisely, and we also exploit the relationships between quadrics mathematics and geometry to improve robustness.
\section{Quadrics}
Before delving into the methodological details, we first define the \textit{quadrics}, a parametric representation of surfaces that is more exhaustive in terms of primitive types and more concise with respect to the mathematical form than existing representations \cite{DBLP:conf/cvpr/LiSDYG19,DBLP:conf/eccv/SharmaLMKCM20,tulsiani2017learning,paschalidou2019superquadrics}. We further decompose the formulation of quadrics to reveal the relationships between their mathematics and geometric attributes (\textit{type}, \textit{scale} and \textit{pose}). Finally, we explain the \textit{degeneracy} of quadrics.

\subsection{Quadrics Representation}
In algebraic geometry, quadrics are a class of surfaces defined implicitly by a second-degree polynomial equation:
\begin{equation}
\begin{aligned}
  f(\mathbf{x,q})&=Ax^2+By^2+Cz^2+2Dxy+2Exz\\
  &+2Fyz+2Gx+2Hy+2Iz+J=0,
\end{aligned}
\label{eq: quadrics equation}
\end{equation}
where $\mathbf{x}$ is a point with homogeneous coordinate $[x,y,z,1]^{\mathrm{T}}$, $\mathbf{q}=[A,B,C,D,E,F,G,H,I,J]$ and the quadratic term is not all zero. The compact matrix form is $\mathbf{x}^{\mathrm{T}}\mathbf{Q}\mathbf{x}=0$, where
\begin{equation}
\setlength{\arraycolsep}{4.5pt}
\mathbf{Q}=\left[\begin{array}{@{}cccc@{}}
A & D & E & G \\
D & B & F & H \\
E & F & C & I \\
G & H & I & J
\end{array}\right],\nabla\mathbf{Q}=2\left[\begin{array}{@{}cccc@{}}
A & D & E & G \\
D & B & F & H \\
E & F & C & I
\end{array}\right].
\label{eq: Q and normal}
\end{equation}
The gradient at $\mathbf{x}$ is $\nabla f(\mathbf{x,q})=\nabla\mathbf{Q}\mathbf{x}$, and also the direction of its normal $\mathbf{n} \in \mathbb{R}^{3} $.

Despite having only 10 parameters, quadrics can uniformly represent 17 geometric primitives, including points, lines, planes, spheres, cylinders, and cones, which encompass most cases in artificial structures.
\subsection{Mathematical Decomposition of Quadrics}
For any quadric $\mathbf{Q}$ in space, all its axes can be aligned to the coordinate axes by precisely applying rotation and translation. After that, $\mathbf{Q}$ is reduced to a diagonal matrix $\mathbf{C}$, namely the \textit{canonical matrix} of a quadric. The canonical matrices of typical quadrics are summarized in Table \ref{tab: Characteristics of Important Quadrics}. It is worth noting that the form of $\mathbf{C}$ and value of $\lambda$ in $\mathbf{C}$ determine respectively the \textit{type} and \textit{scale} of a quadric, and $\mathbf{I}_{\mathbf{s,R,t}} \in \{0,1\}^{3}$ indicates the degeneracy of the \textit{scale}, \textit{rotation}, and \textit{translation}.

\begin{table}[h]
	\centering
	\caption{Characteristics of typical quadrics}
	\label{tab: Characteristics of Important Quadrics}
	\begin{tabular}{llccc}
		\toprule
		Type & $\mathrm{Diag}(\mathbf{C})$ & $\mathbf{I_{s}}$ & $\mathbf{I_{R}}$& $\mathbf{I_{t}}$  \\ \midrule
        Line & $[\lambda_a, \lambda_b, 0, 0]$ & $[0, 0, 0]$ & $[0, 0, 1]$ & $[1, 1, 0]$\\
        Plane & $[\lambda_a, 0, 0, 0]$ & $[0, 0, 0]$ & $[1, 0, 0]$ & $[1, 0, 0]$\\
        Sphere & $[\lambda_a, \lambda_b, \lambda_c, -1]$ & $[1, 1, 1]$ & $[0, 0, 0]$ & $[1, 1, 1]$\\
        Cylinder & $[\lambda_a, \lambda_b, 0, -1]$ & $[1, 1, 0]$ & $[0, 0, 1]$ & $[1, 1, 0]$\\
        Cone & $[\lambda_a, \lambda_b, -\lambda_c, 0]$ & $[1, 1, 0]$ & $[0, 0, 1]$ & $[1, 1, 1]$\\ 
        \bottomrule
	\end{tabular}
\end{table}

\begin{figure}[t]
  \centering
   \includegraphics[width=0.6\linewidth]{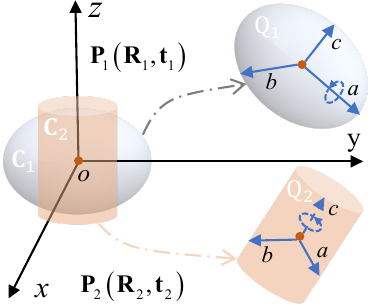}
   \caption{Illustration of quadrics derivation and degeneracy. Ellipsoid $\mathbf{Q}_{1}$ and cylinder $\mathbf{Q}_{2}$ are derived from the canonical quadric $\mathbf{C}_{1}$ and $\mathbf{C}_{2}$ through transformations $\mathbf{P}_{1}$ and $\mathbf{P}_{2}$. For $\mathbf{Q}_{1}$, the \textit{rotation} around axis $a$ is degenerate because the ellipsoid is symmetric on the axes $b$ and $c$ $(\mathbf{I_{R}}=[1,0,0]$). For $\mathbf{Q}_{2}$, the \textit{translation} along axis $c$ and the \textit{rotation} around it are degenerate because the cylinder is open on axis $c$ $(\mathbf{I_{t}}=[1,1,0])$ and symmetric on the axes $a$ and $b$ ($\mathbf{I_{R}}=[0,0,1]$).}
   \label{fig:3-1}
\end{figure}

Inversely, as illustrated in Fig. \ref{fig:3-1}, a given quadric $\mathbf{Q}$ can be regarded as being transformed from $\mathbf{C}$ by
\begin{equation}
\mathbf{Q}=\mathbf{P}^{-\mathrm{T}}\mathbf{C}\mathbf{P}^{-1} = \left[\begin{array}{@{}cc@{}}
\mathbf{R} & \mathbf{t} \\
\mathbf{0}^{\mathrm{T}} & 1
\end{array}\right]^{-\mathrm{T}} \left[\begin{array}{@{}cc@{}}
\mathbf{\Lambda} & \mathbf{0} \\
\mathbf{0}^{\mathrm{T}} & c_{44}
\end{array}\right] \left[\begin{array}{@{}cc@{}}
\mathbf{R} & \mathbf{t} \\
\mathbf{0}^{\mathrm{T}} & 1
\end{array}\right]^{-1}.
\label{eq: Q=PCP}
\end{equation}
where $\mathbf{P}(\mathbf{R},\mathbf{t})\in SE(3)$ denotes the transform matrix from $\mathbf{C}$ to $\mathbf{Q}$, namely the \textit{pose} of a quadric. $\mathbf{R} \in SO(3)$ and $\mathbf{t} \in \mathbb{R}^{3}$ are the rotation and translation blocks of $\mathbf{P}$, $\mathbf{\Lambda} \in \mathbb{R}^{3\times 3}$ and $c_{44}$ are the diagonal blocks of $\mathbf{C}$. Furthermore, $\mathbf{Q}$ can be decomposed as
\begin{equation}\label{eq: quadrics further expanded}
\mathbf{Q} 
 =\left[\begin{array}{@{}cc@{}}
\mathbf{R} \mathbf{\Lambda} \mathbf{R}^{\mathrm{T}} & -\mathbf{R} \mathbf{\Lambda} \mathbf{R}^{\mathrm{T}}\mathbf{t} \\
* & k
\end{array}\right]=\left[\begin{array}{@{}cc@{}}
\mathbf{Q}_{33} & \mathbf{l} \\
* & k
\end{array}\right].
\end{equation}
Here, it is obvious that the \textit{scale} and \textit{pose} of a quadric can be inferred by mathematical analysis of $\mathbf{Q}_{33} \in \mathbb{R}^{3\times 3}$ and $\mathbf{l}\in \mathbb{R}^{3}$, which is explained in Sect. \ref{Loss Functions}.

\subsection{Geometric Degeneracy of Quadrics}
\label{Degeneracy of Quadrics}
The degeneracy means that the \textit{scale}, \textit{rotation}, and \textit{translation} of a quadric won’t affect its shape, which can be judged from $\mathbf{Q}$. If $\mathbf{Q}$ is rank-deficient, this will lead to the degeneracy of \textit{scale} at certain axes. Since $\mathbf{Q}_{33}$ is a real symmetric matrix, it can be similarly diagonalized as $\mathbf{Q}_{33} =\hat{\mathbf{R}}  \hat{\mathbf{\Lambda } } \hat{\mathbf{R}} ^{\mathrm{T}}$, where $\hat{\mathbf{\Lambda}} = \mathbf{\Lambda}$ can be denoted by the eigenvalues of $\mathbf{Q}_{33}$ as $\mathrm{diag}\left ( \lambda_a,\lambda_b,\lambda_c  \right )$ and $\hat{\mathbf{R}}$ consists of the eigenvectors.  The zero values in the eigenvalues make the \textit{translation} along the corresponding axis degenerate. If there are duplicates in the eigenvalues, \textit{i.e.}, the quadric is symmetric, and the \textit{rotation} around the non-symmetric axis will degenerate. 
Fig. \ref{fig:3-1} illustrates this more intuitively.

\section{Methodology}

\begin{figure*}
  \centering
  \includegraphics[width=0.8\linewidth]{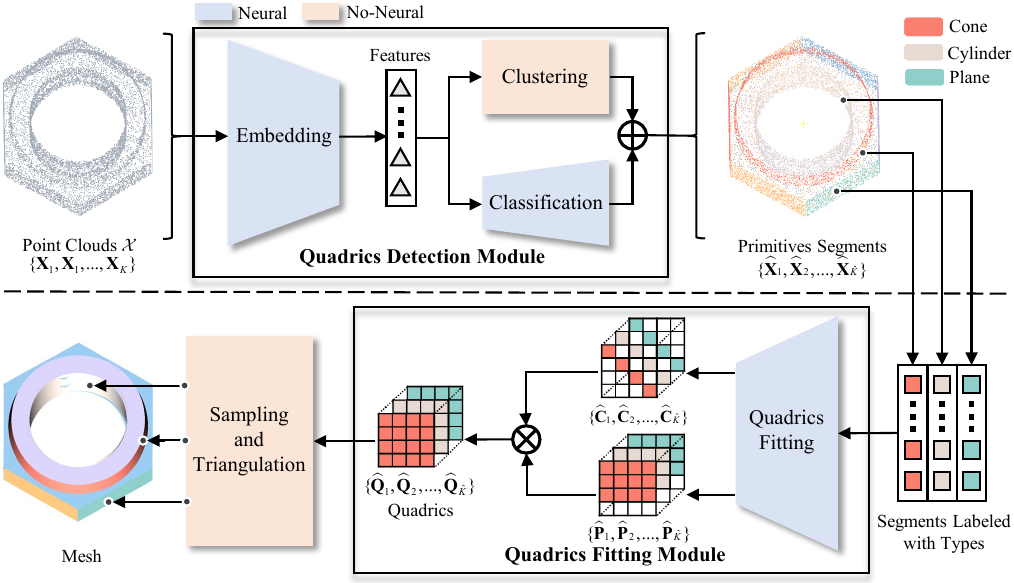}
  \caption{Overview of QuadricsNet. The detection module segments point clouds into several quadrics segments and identifies the quadrics types, then the fitting module fits the quadrics parameters of each segment, and finally the structure map is reconstructed using the fitted quadrics.}
  \label{fig: Overview of QuadricsNet}
\end{figure*}

We presume that the inputs are point clouds $\mathcal{X}\in  \mathbb{R}^{N}$ with 3D positions and optional normals. We seek to represent it with a set of quadrics $\{\hat{\mathbf{Q}}_1,\hat{\mathbf{Q}}_2,...,\hat{\mathbf{Q}}_{\hat{K}}\}$ that closely approximate its underlying surfaces. For this purpose, an end-to-end framework \textit{QuadricsNet} is proposed, as illustrated in Fig. \ref{fig: Overview of QuadricsNet}. We first introduce the two modules of \textit{QuadricsNet}: \textit{quadrics detection module} and \textit{quadrics fitting module}, and then we specify the losses customized for quadrics.

\subsection{Quadrics Detection Module}
The purpose of this module is to decompose the input $\mathcal{X}$ into several quadrics segments $\{\hat{\mathbf{X}}_1,\hat{\mathbf{X}}_2,...,\hat{\mathbf{X}}_{\hat{K}}\}$ while simultaneously identifying their quadrics types $\{\hat{l}_1,\hat{l}_2,...,\hat{l}_{\hat{K}}\}$. To this end, we implement it in three steps: \textit{embedding}, \textit{clustering}, and \textit{classification}.

\textbf{Embedding.} We primarily employ an embedding network to learn the point-wise features $\mathbf{Z} \in  \mathbb{R}^{N\times128}$ that can distinguish different quadrics segments in $\mathcal{X}$. Its backbone is derived from EdgeConv \cite{DBLP:journals/tog/WangSLSBS19}, which can learn local features singly and global features when stacking in multiple.

\textbf{Clustering.} Based on the features $\mathbf{Z}$ with implicit distinguishability, we proceed to modify them by the mean shift \cite{DBLP:journals/tit/FukunagaH75} in a differentiable way to explicitly cluster the points according to the quadrics segments:
\begin{equation}
\mathbf{Z}_{t+1} = \mathbf{Z}_{t}+\eta (\mathbf{Z}_{t}\mathbf{KD}^{-1}-\mathbf{Z}_{t}),
\label{mean shift}
\end{equation}
where $\mathbf{K}$ is a Gaussian kernel, $\mathbf{D}=\sum\mathbf{K}$ and the step size $\eta = 1$. After the convergence of Eq. \ref{mean shift}, the cluster centers are determined by a non-maximal suppression. Eventually, each point is assigned to a segment according to its nearest center. We employ a membership matrix $\hat{\mathbf{W}}\in \{0,1\}^{N \times \hat{K}}$ to indicate the affiliations between points $\mathcal{X}$ and clustered segments  $\{\hat{\mathbf{X}}_1,\hat{\mathbf{X}}_2,...,\hat{\mathbf{X}}_{\hat{K}}\}$ , where we ensure that $\hat{K}\ge K$ even if $K$ varies for different objects.



\textbf{Classification.} Simultaneously with clustering, we combine an MLP and a softmax to construct primitive classifiers that predict the point-wise quadrics type (\textit{e.g.}, plane, sphere, cylinder, cone) based on $\mathbf{Z}$. We use another membership matrix $\hat{\mathbf{L}}\in [0,1]^{N \times L}$ to indicate the probability of each point in $\mathcal{X}$ being predicted as which quadrics type. Eventually, the types $\{\hat{l}_1,\hat{l}_2,...,\hat{l}_{\hat{K}}\}$ of clustered segments are determined by a majority voting overall types of points in each segment.

\subsection{Quadrics Fitting Module}
\label{Quadrics Fitting Module}
This module fits quadrics parameters $\{\hat{\mathbf{Q}}_1,\hat{\mathbf{Q}}_2,...,\hat{\mathbf{Q}}_{\hat{K}}\}$ for all detected quadrics segments $\{\hat{\mathbf{X}}_1,\hat{\mathbf{X}}_2,...,\hat{\mathbf{X}}_{\hat{K}}\}$. 

Prior to fitting, the points $\hat{\mathbf{X}}_k$ of each segment are derived by filtering $\mathcal{X}$ according to $\hat{\mathbf{W}}$:
\begin{equation}
\hat{\mathbf{X}}_k = \hat{\mathbf{W}}_{:,k} \odot \mathcal{X},
\end{equation}
where $\odot$ denotes the element-wise multiplication. We construct a quadrics fitting network to robustly estimate the quadrics parameters $\hat{\mathbf{Q}}_{k}$ for the underlying surface of the segment $\hat{\mathbf{X}}_{k}$. The backbone of the fitting network is similar to the embedding network. Furthermore, we leverage two MLPs to respectively estimate a \textit{canonical matrix} $\hat{\mathbf{C}}_{k}$ and a \textit{inverse pose matrix} $\hat{\mathbf{P}}_{k}^{-1}(\hat{\mathbf{R}}_k,\hat{\mathbf{t}}_k)$, which correspond to the \textit{scale} and \textit{pose} respectively. According to Eq. \ref{eq: Q=PCP}, the quadric is finally determined as
\begin{equation}
\hat{\mathbf{Q}}_{k}=\hat{\mathbf{P}}_{k}^{-\mathrm{T}} \hat{\mathbf{C}}_{k} \hat{\mathbf{P}}_{k}^{-1}.
\label{eq: Q=PCP fitting}
\end{equation}

It is worth noting that depending on the symmetry and sparsity of $\hat{\mathbf{C}}_{k}$ and $\hat{\mathbf{P}}_{k}^{-1}$, the number of parameters in the output of the fitting network is correspondingly no more than 10 and 12. According to Table \ref{tab: Characteristics of Important Quadrics}, the form of $\hat{\mathbf{C}}_{k}$ determines the \textit{type} of a quadric. To guarantee that the fitted $\hat{\mathbf{Q}}_{k}$ is consistent with the type $\hat{l}_{k}$, we leverage $\hat{l}_{k}$ as a prior to constrain the form of output $\hat{\mathbf{C}}_{k}$, \textit{e.g.}, the form is constrained to be $\mathrm{diag}( [ \hat{\lambda}_a, \hat{\lambda}_b, 0, -1 ])$ if a cylinder is to be fitted. In such a concise representation, even a small parameter error can lead to the wrong fitted type. Our network avoids this issue and significantly improves the fitting accuracy.

\subsection{Losses}
\label{Loss Functions}
Six losses customized for quadrics are defined from mathematic and geometric perspectives to train these modules.

\textbf{Losses for Detection.} We adopt the GT segment membership $\mathbf{W}$ and type membership $\mathbf{L}$ to supervise this module.

We leverage the \textit{triplet loss} \cite{DBLP:journals/corr/HermansBL17} to achieve the point-wise features $\mathbf{Z}$ with segment distinguishabilities:
\begin{equation}
\mathcal{L}_\mathrm{emb} = \frac{1}{M} \sum_{i=1}^{M} \max \left ( \left \|  \mathbf{z}^\mathrm{A}_i -\mathbf{z}^\mathrm{P}_i \right \|^2 - \left \|  \mathbf{z}^\mathrm{A}_i -\mathbf{z}^\mathrm{N}_i \right \|^2 + \alpha  ,0  \right )  ,
\end{equation}
where $\mathbf{z} \in  \mathbb{R}^{128}$ is the point-wise feature, $\alpha$ is the margin, $M$ is the total number of triplet sets $\{\mathbf{z}^\mathrm{A},\mathbf{z}^\mathrm{P},\mathbf{z}^\mathrm{N}\}$ sampled from features of different segments according to $\mathbf{W}$. The \textit{cross entropy loss} $H$ is employed for quadrics classification:
\begin{equation}
\mathcal{L}_\mathrm{type} = \frac{1}{N} \sum_{i=1}^{N} H\left ( \hat{\mathbf{L}}_{i,:} ,\mathbf{L}_{i,:} \right ).
\end{equation}

\textbf{Losses for Fitting.} To supervise the fitting module, we employ the GT $\mathbf{Q}$ and normals $\mathbf{N}$.

Previously, to match the GT signals to the fitted signals of each clustered segment, we first compute the \textit{Relaxed Intersection over Union} of $\hat{\mathbf{W}}_{:,k}$ and $\mathbf{W}_{:,k}$, then the best one-to-one correspondence between GT segments $\mathbf{X}_{k}$ and clustered segments $\mathbf{\hat{X}}_{k}$ is matched by the \textit{Hungarian algorithm} \cite{https://doi.org/10.1002/nav.3800020109}. According to Eq. \ref{eq: quadrics equation}, we define the \textit{primal loss} for $\mathbf{\hat{Q}}$, which roughly mirrors the quality of fitting:
\begin{equation}
\mathcal{L}_\mathrm{primal}=\frac{1}{\hat{K}} \sum_{k=1}^{\hat{K}}\frac{1}{|\mathbf{\hat{X}}_{k}|}\sum_{\mathbf{\hat{x}}\in\mathbf{\hat{X}}_{k}} \left \| \mathbf{\hat{x}}^{\mathrm{T}}\mathbf{Q}_k\mathbf{\hat{x}}\right \|^2.
\end{equation}
Further, based on the correlation between the normals and parameters of quadrics (Eq. \ref{eq: Q and normal}), we define the \textit{normal loss}:
\begin{equation}
\mathcal{L}_\mathrm{normal}=\frac{1}{\hat{K}} \sum_{k=1}^{\hat{K}}\frac{1}{|\mathbf{\hat{X}}_{k}|}\sum_{\mathbf{\hat{x}}\in\mathbf{\hat{X}}_k}\left \| \nabla\hat{\mathbf{Q}}_k\hat{\mathbf{x}}\otimes  \mathbf{n} \right \|^2,
\end{equation}
where $\otimes$ denotes column-wise cross product, $\mathbf{n}$ is GT normal on $\mathbf{\hat{x}}$. Additionally, we explicitly define \textit{the regression loss}:
\begin{equation}
\mathcal{L}_\mathrm{reg}=\frac{1}{\hat{K}} \sum_{k=1}^{\hat{K}} \left \|\hat{\mathbf{Q}}_k-\mathbf{Q}_k  \right \|^2 .
\end{equation}

The \textit{scale} and \textit{pose} of fitted quadrics can be learned directly in Eq. \ref{eq: Q=PCP fitting}, but those of GT need to be inferred by mathematical analysis of its $\mathbf{Q}$. Prior to inference, $\mathbf{Q}$ has to be normalized to eliminate proportion ambiguity in Eq. \ref{eq: quadrics equation}:
\begin{equation}
{\mathbf{Q}} = \left\{ \begin{array}{cc}
\left| {\dfrac{{\prod {\lambda _i^{\mathbf{Q}_{33}}} }}{{\prod {\lambda _i^\mathbf{Q} }}}} \right|{\mathbf{Q}}, \ &{c_{44}} \ne 0\\
\dfrac{1}{\left \| \mathbf{Q} \right \| }{\mathbf{Q}}, \ &{c_{44}} = 0
\end{array} ,\right.
\end{equation}
where $\lambda^{\mathbf{Q}_{33}}$ and $\lambda^{\mathbf{Q}}$ are the non-zero eigenvalues of $\mathbf{Q}_{33}$ and $\mathbf{Q}$. According to Sect. \ref{Degeneracy of Quadrics}, $\mathbf{Q}_{33} =\tilde{\mathbf{R}}  \mathbf{\Lambda } \tilde{\mathbf{R}} ^{\mathrm{T}}$, where $\mathbf{\Lambda}=\mathrm{diag}\left ( \lambda_a,\lambda_b,\lambda_c  \right )$. Without loss of generality, we assume that $\lambda_a > \lambda_b > \lambda_c$, and the \textit{scale} $\mathbf{s} \in \mathbb{R}^{3} $, \textit{rotation} $\mathbf{R} \in SO(3)$ and \textit{translation} $\mathbf{t} \in \mathbb{R}^{3}$ are
\begin{equation}
\left\{ \begin{array}{lll}
\left [s_a,s_b,s_c\right ]^{\mathrm{T}}&=&\mathrm{diag}(\mathbf{I_s})\sqrt{\left | \left [ \frac{1}{\lambda_a} ,\frac{1}{\lambda_b},\frac{1}{\lambda_c} \right ]\right | },\\
\left [ \mathbf{r}_a,\mathbf{r}_b,\mathbf{r}_c\right ]  &=& \pm ( \mathrm{diag} (\mathbf{I} _\mathbf{R})\tilde{\mathbf{R}}^{\mathrm{T}})^{\mathrm{T}},\\
\left [ t_a,t_b,t_c\right ] ^{\mathrm{T}} &=& \mathrm{diag}(\mathbf{I_t})\tilde{\mathbf{t}},
\end{array} \right.
\end{equation}
where the direction of $\mathbf{r}$ can either be identical or opposite to the column of $\tilde{\mathbf{R}}$ and $\tilde{\mathbf{t}} \in \left \{\tilde{\mathbf{t}}\mid \mathbf{Q}_{33} \tilde{\mathbf{t}} + \mathbf{l}= 0 \right \}$. Based on these theoretical guides, we define \textit{the geometric loss}:

\begin{equation}
\mathcal{L}_\mathrm{geo}=\frac{1}{\hat{K}}\sum_{k=1}^{\hat{K}} \left [ ||\hat{\mathbf{s}}_k-\mathbf{s}_k||^2+||\hat{\mathbf{R}}_k\otimes \mathbf{R}_k||^2+ || \hat{\mathbf{t}}_k-\mathbf{t}_k ||^2 \right ]  ,
\end{equation}
where we approximate them for gradient stabilization:
\begin{equation}
\left\{ \begin{array}{lll}
\hat{\mathbf{s}}-\mathbf{s} 
& \approx&\hat{\mathbf{C}}-\mathbf{\Lambda },\\
\hat{\mathbf{R}}\otimes \mathbf{R} & =&\sum_{i=1}^{3} \hat{\mathbf{R}}_{:,i}\times \mathbf{R}_{:,i},\\
 \hat{\mathbf{t}}-\mathbf{t} & \approx &\mathbf{\Lambda } \mathbf{R} ^{\mathrm{T}} \hat{\mathbf{t}}+\mathbf{R} ^{\mathrm{T}}\mathbf{l}.
\end{array} \right.
\end{equation}



\section{Experiments}
\label{sec:experiments}
In this section, we detail the experimental dataset, evaluation metrics, and result comparisons to the state-of-the-art.  
\subsection{Quadrics Dataset and Training Strategy}
\label{Quadrics Dataset}
Since there is no publicly available point cloud dataset with labeled quadrics parameters $\mathbf{Q}$, we build a \textit{Quadrics Dataset} for \textit{QuadricsNet} training and evaluation using simulated quadric segments and CAD objects selected from ABC dataset \cite{DBLP:conf/cvpr/KochMJWABAZP19}. To improve the network robustness to incomplete point clouds, we randomly trim them to simulate fragmentation and incompletion. Limited by the ABC dataset, we mainly consider four types of quadrics, that is, planes, spheres, cylinders, and cones. We compute $\mathbf{Q}$ for all segments in this dataset and also randomly add noise along the normal direction in a uniform range $[-0.01,0.01]$. It contains two subsets: a) \textit{Segment Dataset} is mainly designed for quadrics fitting network, which has 20k quadrics segments for training and 3k segments for testing; b) \textit{Object Dataset} is built for both quadrics detection and fitting networks, which has 20k CAD objects for training and 3k objects for testing. 


We adopt a two-step pre-training and fine-tuning strategy to train the \textit{QuadricsNet}. We first pre-train the detection network using $\mathcal{L}_\mathrm{emb}+\mathcal{L}_\mathrm{type}$ loss on the \textit{Object Dataset}. At the same time, we pre-train the fitting network with $\mathcal{L}_\mathrm{primal}+\mathcal{L}_\mathrm{normal}+\mathcal{L}_\mathrm{reg}+\mathcal{L}_\mathrm{geo}$ loss on the \textit{Segment Dataset}. Then, the whole \textit{QuadricsNet} is fine-tuned with the six losses on the \textit{Object Dataset} in an end-to-end manner.

\subsection{Evaluation Metrics}
To quantitatively evaluate the performance of \textit{QuadricsNet}, we use Seg-IoU and Type-IoU metrics to measure the performance of quadrics detection, while Residual and P-coverage metrics measure the accuracy of quadrics fitting. 

\begin{itemize}
\item[$\bullet$] \textbf{Seg-IoU} (S-IoU): $\frac{1}{K}{\textstyle \sum_{k=1}^{K}} IoU(\mathbf{\hat{W}}_{:,k},\mathbf{W}_{:,k})$ measures the accuracy of quadrics segment clustering.
\item[$\bullet$] \textbf{Type-IoU} (T-IoU): $\frac{1}{K}{\textstyle \sum_{k=1}^{K}} \mathbb{I}(\hat{l}_k=l_k) $ measures the accuracy of quadrics classification.
\item[$\bullet$] \textbf{Residual} (Res): $\frac{1}{K}{\textstyle \sum_{k=1}^{K}} \frac{1}{|\mathbf{X}_k|} {\textstyle\sum_{\mathbf{x}\in \mathbf{X}_k}} D(\mathbf{\hat{Q}}_k,\mathbf{x})$ is the average Euclidean distance of raw points $\mathbf{x}$ to the predicted quadric surfaces.
\item[$\bullet$] \textbf{P-coverage} (P-cov): $\frac{1}{|\mathcal{X}|}{\textstyle \sum_{\mathbf{x}\in \mathcal{X}}} \mathbb{I}[ \mathrm{min}_{k=1}^{K}D(\mathbf{\hat{Q}}_k,\mathbf{x})<\epsilon ]$ measures the percentage of input points covered by the predicted quadric surfaces.
\end{itemize}

\subsection{Result Comparisons}
To evaluate the performance, we compare \textit{QuadricsNet} with five state-of-the-art geometric primitive parsing methods, including the traditional nearest neighbor (NN) \cite{sharma2018csgnet} and Efficient RANSAC \cite{schnabel2007efficient}, the learning-based SPFN \cite{DBLP:conf/cvpr/KochMJWABAZP19}, ParseNet \cite{DBLP:conf/eccv/SharmaLMKCM20} and HPNet \cite{yan2021hpnet}. Unlike our unified quadrics representation for different geometric primitives, these competitors employ individual representations for them. All these methods are evaluated on the \textit{Object Dataset} test set for a fair comparison.

\textbf{Quantitative Results.} As reported in Table \ref{tab: Quantitative comparison}, we test these methods with two input cases: points (p) and points with normals (p+n). Under the same input information, \textit{QuadricsNet} generally outperforms other methods for all metrics. Especially for the Residual and P-coverage metrics, our unified quadrics representation yields better performance than other non-unified methods with different representations for each primitive. These results demonstrate the effectiveness of our quadrics-based framework on geometric primitive detection and fitting.

\begin{figure*}
  \centering
  \includegraphics[width=\linewidth]{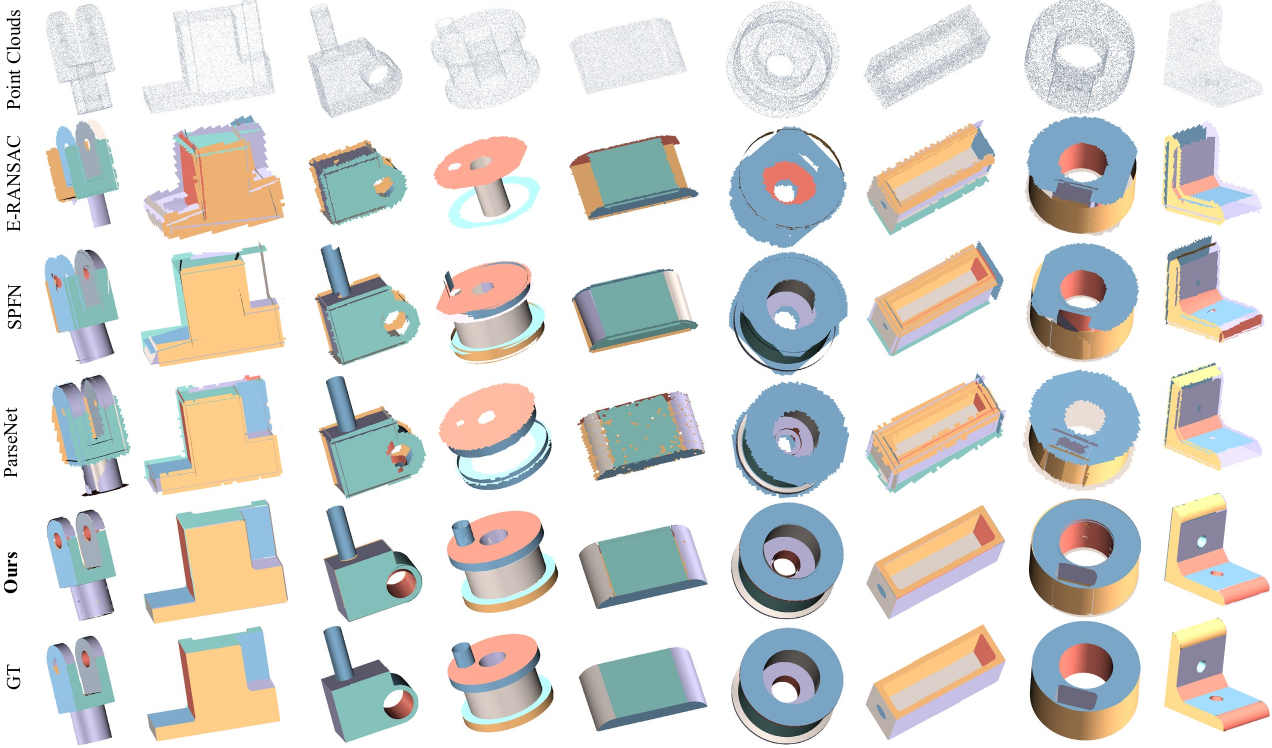}
  \caption{Qualitative comparisons. Structure mapping of the raw point clouds using different primitive parsing methods.}
  \label{fig: Qualitative Comparisons.}
\end{figure*}


\begin{table}[h]\scriptsize
    \centering
    \setlength{\tabcolsep}{2.5pt}
    \caption{Quantitative comparisons}
    \label{tab: Quantitative comparison}
    \begin{tabular}[t]{l|c|c|c|c|cc}
        \toprule
        \multirow{2}{*}{Method} & \multirow{2}{*}{Input} & \multirow{2}{*}{S-IoU$(\%)$ $\uparrow$} & \multirow{2}{*}{T-IoU$(\%)$ $\uparrow$} & \multirow{2}{*}{Res $\downarrow$} & \multicolumn{2}{c}{P-cov $(\%)$ $\uparrow$} \\
        \cline{6-7}
        & & & & & $\epsilon=0.01$ & $\epsilon=0.02$ \\
        \midrule
        NN \cite{sharma2018csgnet} & p & 54.10 & 61.10 & - & - & - \\
        \midrule
        E-RANSAC \cite{schnabel2007efficient} & p+n & 67.21 & - & 0.022 & 83.40 & 87.73 \\
        \midrule
        \multirow{2}{*}{SPFN \cite{DBLP:conf/cvpr/KochMJWABAZP19}}  & p & 61.42 & 74.56 & 0.023 & 82.55 & 90.67 \\
        & p+n & 73.19 & 85.91 & 0.019 & 86.79 & 92.14 \\
        \midrule
        \multirow{2}{*}{ParseNet \cite{DBLP:conf/eccv/SharmaLMKCM20}} & p & 74.12 & 79.90 & 0.018 & 83.20 & 92.32 \\
        & p+n & 85.70 & 90.21 & 0.013 & 89.76 & 93.98 \\
        \midrule
        \multirow{2}{*}{HPNet \cite{yan2021hpnet}} & p & 80.32 & 87.19 & 0.014 & 86.07 & 94.12 \\
        & p+n & 88.17 & 92.25 & \textbf{0.009} & 93.12 & 96.27 \\
        \midrule
        \multirow{2}{*}{Ours} & p & \textbf{84.12} & \textbf{88.00} & \textbf{0.013} & \textbf{88.64} & \textbf{95.88}\\
        & p+n & \textbf{92.16} & \textbf{95.87} & \textbf{0.009} & \textbf{93.76} & \textbf{97.12}\\
        \bottomrule
    \end{tabular}\vspace{-0mm}
\end{table}

\begin{figure}
  \centering
  \includegraphics[width=1\linewidth]{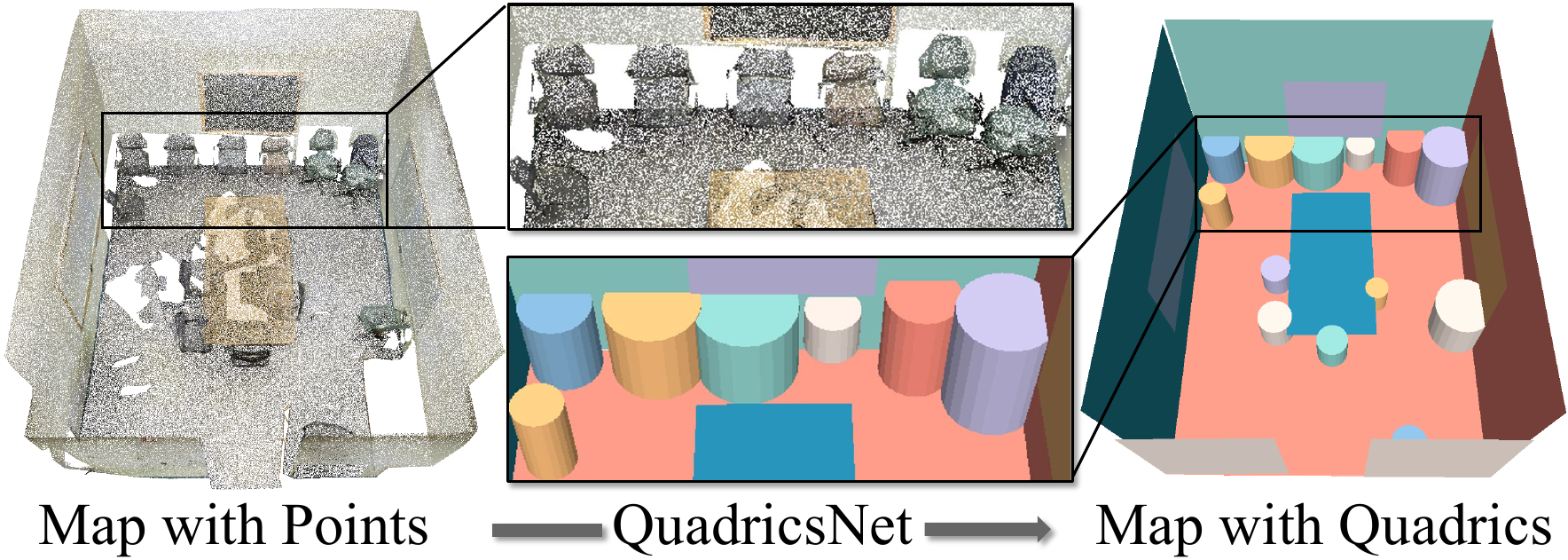}
  \caption{Stucture mapping of a scene in S3DIS using quadrics.}
  \label{fig: Generalization Tests.}\vspace{-1.0mm}
\end{figure}

\textbf{Qualitative Results.} 
Fig. \ref{fig: Qualitative Comparisons.} qualitatively shows the structure mapping of point clouds using different primitive parsing methods. The mapping results using QuadricsNet are more reasonable with clean boundaries and better structure integrity because QuadricsNet can detect and fit primitives more precisely. Furthermore, to evaluate the generalizability of \textit{QuadricsNet} in real-world data, we extend the experiment to the large-scale indoor S3DIS dataset \cite{armeni_cvpr16}. As shown in Fig. \ref{fig: Generalization Tests.}, our method effectively represents the real scene with quadrics on the object-level scale and yields a robust structure mapping result.

\begin{table}[h]
    \centering
    \caption{Impacts of loss design on quadrics fitting}
    \label{tab:Ablation Study}
    \begin{tabular}[t]{cccc|c|cc}
        \toprule
        \multicolumn{4}{m{2.5cm}|}{Loss settings} & \multirow{2}{*}{Res $\downarrow$} & \multicolumn{2}{c}{P-cov $(\%)$ $\uparrow$} \\
        \cline{1-4}
        \cline{6-7}
        $\mathcal{L}_\mathrm{primal}$ & $\mathcal{L}_\mathrm{normal}$ & $\mathcal{L}_\mathrm{reg}$ & $\mathcal{L}_\mathrm{geo}$ &  & $\epsilon=0.01$ & $\epsilon=0.02$  \\
        \midrule
        \checkmark &  &  &  & 0.032 & 68.10 & 71.87 \\
        \checkmark & \checkmark &  &  & 0.029 & 73.03  & 78.62 \\
        \checkmark & \checkmark & \checkmark &  & 0.012 & 89.45  & 90.34 \\
        \checkmark & \checkmark & \checkmark & \checkmark & \textbf{0.008} & \textbf{94.12}  & \textbf{96.10} \\
        \bottomrule
    \end{tabular}
\end{table}

\subsection{Ablation Study}
We mainly discuss the effectiveness of the designed loss functions on the quadrics fitting module. By gradually adding the four losses, we obtain their impacts in Table \ref{tab:Ablation Study}. On the basis of the $\mathcal{L}_\mathrm{primal}$ loss, adding other supervision terms generally improves the fitting accuracy. Especially for adding the $\mathcal{L}_\mathrm{reg}$ loss, we obtain a noticeable performance improvement with a large margin. Eventually, adding the $\mathcal{L}_\mathrm{geo}$ loss produces the best result, which demonstrates the effectiveness of combining mathematical factors of quadrics and geometric attributes on geometric primitive parsing.

\section{Conclusions and Future Work}
\label{sec:conclusion}
In this paper, we propose an end-to-end \textit{QuadricsNet} to learn a concise representation of geometric primitives in point clouds. \textit{Quadrics} representation successfully unifies different geometric primitives, while the geometric attributes from quadrics mathematical formulation effectively supervise the quadrics detection and fitting networks. Experiments of primitive parsing on the collected dataset and structure mapping on real-world scenes demonstrate that our quadrics representation is effective and the \textit{QuadricsNet} framework is robust. In the future, we will explore the fusion of geometry and semantics for primitive parsing and structure mapping.  

\section{Acknowledgement}
This work was supported by the NSFC grants under contracts Nos. 62301370 and 62325111, and Wuhan University-Huawei Geoinformatics Innovation Laboratory.

\bibliographystyle{IEEEtran.bst}
\bibliography{references}

\end{document}


\maketitle

\section{Notation}
Following the notation convention of the submitted paper, the quadric states in the world frame is represented by $(\mathbf{R}_q, \mathbf{t}_q, \mathbf{s}_q)$, and the robot pose is $(\mathbf{R}_r, \mathbf{t}_r)$. Therefore, the state vector involved in a single observation is $\mathbf{x} = [\mathbf{R}_r, \mathbf{t}_r, \mathbf{R}_q, \mathbf{t}_q, \mathbf{s}_q]$.

To simplify the presentation, we derive the Jacobian matrix based on a single observation, while the complete Jacobian can be constructed by filling in per observation Jacobians. 

\section{Error Function}
As presented in the paper, the observation error is given by
\begin{equation}
    \mathbf{e} = \begin{pmatrix}
        \mathbf{e}_{\mathbf{R}} \\ 
        \mathbf{e}_{\mathbf{t}} \\
        \mathbf{e}_{\mathbf{s}}
    \end{pmatrix} =\left(\begin{array}{l}
    \text{diag}(\mathbf{I^R}) \left(\mathbf{V} \otimes \mathbf{\Delta_R}\right)^T\\
    \text{diag}(\mathbf{I}^{\mathbf{t}}) (\mathbf{DV}^T\mathbf{\Delta_t+V}^T\mathbf{l})\\
    \text{diag}(\mathbf{I}^{\mathbf{s}}) (\mathbf{s}^2 - \mathbf{\Lambda})
    \end{array}\right)\in \mathbb{R}^{15}
\end{equation}
where $\otimes$ means column-wise cross product. $\mathbf{\Delta}_\mathbf{R} = \mathbf{R}_r^T\mathbf{R}_q$ and $\mathbf{\Delta_t} = \mathbf{R}_r^T(\mathbf{t}_q - \mathbf{t}_r)$ are the rotation and translation of quadrics pose transformed into the robot frame. $\mathbf{\Lambda}=[\lambda_1,\lambda_2,\lambda_3]$ is the vector of eigenvalues which are stored as the diagonal elements of $\mathbf{D}$. Here, $\mathbf{e}_{\mathbf{R}}$ is a $3\times 3$ matrix and will be vectorized and then stacked into the error vector. 

Then we take derivative of $\mathbf{e}$ over $\mathbf{x}$: 
\begin{equation}
    \dfrac{\partial \mathbf{e}}{\partial \mathbf{x}} = \begin{pmatrix} \dfrac{\partial \mathbf{e}_\mathbf{R}}{\partial \mathbf{R}_r} & \dfrac{\partial \mathbf{e}_\mathbf{R}}{\partial \mathbf{t}} & \dfrac{\partial \mathbf{e}_\mathbf{R}}{\partial \mathbf{R}_q} & \dfrac{\partial \mathbf{e}_\mathbf{R}}{\partial \mathbf{t}_q} & \dfrac{\partial \mathbf{e}_\mathbf{R}}{\partial \mathbf{s}_q}\\
    \dfrac{\partial \mathbf{e}_\mathbf{t}}{\partial \mathbf{R}_r} & \dfrac{\partial \mathbf{e}_\mathbf{t}}{\partial \mathbf{t}} & \dfrac{\partial \mathbf{e}_\mathbf{t}}{\partial \mathbf{R}_q} & \dfrac{\partial \mathbf{e}_\mathbf{t}}{\partial \mathbf{t}_q} & \dfrac{\partial \mathbf{e}_\mathbf{t}}{\partial \mathbf{s}_q} \\
    \dfrac{\partial \mathbf{e}_\mathbf{s}}{\partial \mathbf{R}_r} & \dfrac{\partial \mathbf{e}_\mathbf{s}}{\partial \mathbf{t}} & \dfrac{\partial \mathbf{e}_\mathbf{t}}{\partial \mathbf{R}_q} & \dfrac{\partial \mathbf{e}_\mathbf{s}}{\partial \mathbf{t}_q} & \dfrac{\partial \mathbf{e}_\mathbf{s}}{\partial \mathbf{s}_q}
    \end{pmatrix} \in \mathbb{R}^{15\times15}
\end{equation}
Note that the first dimension size 15 is the number of constraints or the error terms. The above Jacobian can be simplified by identifying zero blocks: 
\begin{equation}
    \dfrac{\partial \mathbf{e}}{\partial \mathbf{x}} = \begin{pmatrix} \dfrac{\partial \mathbf{e}_\mathbf{R}}{\partial \mathbf{R}_r} & \mathbf{0} & \dfrac{\partial \mathbf{e}_\mathbf{R}}{\partial \mathbf{R}_q} & \mathbf{0} & \mathbf{0}\\
    \dfrac{\partial \mathbf{e}_\mathbf{t}}{\partial \mathbf{R}_r} & \dfrac{\partial \mathbf{e}_\mathbf{t}}{\partial \mathbf{t}} & \dfrac{\partial \mathbf{e}_\mathbf{t}}{\partial \mathbf{R}_q} & \dfrac{\partial \mathbf{e}_\mathbf{t}}{\partial \mathbf{t}_q} & \mathbf{0} \\
    \mathbf{0} & \mathbf{0} & \mathbf{0} & \mathbf{0} & \dfrac{\partial \mathbf{e}_\mathbf{s}}{\partial \mathbf{s}_q}
    \end{pmatrix} \in \mathbb{R}^{15\times15}
    \label{eqn:jacobian}
\end{equation}

Now we rewrite error terms explicitly to prepare for the derivation of $\dfrac{\partial \mathbf{e}}{\partial\mathbf{x}}$:
\begin{equation}
    \begin{aligned}
    \mathbf{e}_{\mathbf{R}} &= \begin{pmatrix} \cdots \\ [\mathbf{v}_i]_{\times}\mathbf{R}_r^T\mathbf{R}_q\mathbf{u}_i\\ \cdots \end{pmatrix} \in \mathbb{R}^{9\times 1} \\
    \mathbf{e}_{\mathbf{t}} &= \begin{pmatrix} \cdots \\ \lambda_i\mathbf{u}_i^T\mathbf{R}_q^T(\mathbf{t}_q-\mathbf{t}) +\mathbf{u}_i^T\mathbf{R}_q^T\mathbf{R}_r\mathbf{l} \\ \cdots \end{pmatrix} \in \mathbb{R}^{3\times 1}\\
    \mathbf{e}_{\mathbf{s}} &= \begin{pmatrix} \cdots \\ \mathbf{s}_i^2 - \lambda_i\\ \cdots \end{pmatrix} \in \mathbb{R}^{3\times 1} 
    \end{aligned}
\end{equation}
where $\mathbf{u}_i$ are unit vectors: 
\begin{equation}
    \mathbf{u}_1 = (1,0,0)^T, \;\mathbf{u}_2 = (0,1,0)^T, \;\mathbf{u}_3 = (0,0,1)^T
\end{equation}
\section{Linearization}
Computing Jacobian involving $\mathbf{R}_r$ and $\mathbf{R}_q$ requires linearization which can be achieved by applying the small angle approximation:
\begin{equation}
    \mathbf{R}_r = \mathbf{R}_r\delta \mathbf{R},\quad \delta\mathbf{R}_r\approx \mathbf{I} + [\mathbf{w}_r]_{\times}
\end{equation}
and
\begin{equation}
    \mathbf{R}_q = \mathbf{R}_q\delta \mathbf{R}_q,\quad \delta\mathbf{R}_q\approx \mathbf{I} + [\mathbf{w}_q]_{\times}
\end{equation}
where $[\cdot]_{\times}$ means skew-symmetric operator and $\mathbf{I}$ is the identity matrix. Now the linearized error function becomes
\begin{equation}
    \begin{aligned}
    \Bar{\mathbf{e}}_\mathbf{R} \lvert_{\mathbf{R}_r}
    &= \begin{pmatrix} \cdots \\ [\mathbf{v}_i]_{\times}(\mathbf{I}+[\mathbf{w}_r]_{\times})^T\mathbf{R}_r^T\mathbf{R}_q\mathbf{u}_i\\ \cdots \end{pmatrix}\\
    &= \begin{pmatrix} \cdots \\ [\mathbf{v}_i]_{\times}(\mathbf{I}-[\mathbf{w}_r]_{\times})\mathbf{R}_r^T\mathbf{R}_q\mathbf{u}_i\\ \cdots \end{pmatrix}\\
    &= \begin{pmatrix} \cdots \\ [\mathbf{v}_i]_{\times}\mathbf{R}_r^T\mathbf{R}_q\mathbf{u}_i - [\mathbf{v}_i]_{\times}[\mathbf{w}_r]_{\times}\mathbf{R}_r^T\mathbf{R}_q\mathbf{u}_i\\ \cdots \end{pmatrix}\\
    &= \begin{pmatrix} \cdots \\ [\mathbf{v}_i]_{\times}\mathbf{R}_r^T\mathbf{R}_q\mathbf{u}_i + [\mathbf{v}_i]_{\times}[\mathbf{R}_r^T\mathbf{R}_q\mathbf{u}_i]_{\times}\mathbf{w}_r\\ \cdots \end{pmatrix}\\
    \Bar{\mathbf{e}}_\mathbf{R} \lvert_{\mathbf{R}_q}
    &= \begin{pmatrix} \cdots \\ [\mathbf{v}_i]_{\times}\mathbf{R}_r^T\mathbf{R}_q(\mathbf{I}+[\mathbf{w}_q]_{\times})\mathbf{u}_i\\ \cdots \end{pmatrix}\\
    &= \begin{pmatrix} \cdots \\ [\mathbf{v}_i]_{\times}\mathbf{R}_r^T\mathbf{R}_q\mathbf{u}_i+[\mathbf{v}_i]_{\times}\mathbf{R}_r^T\mathbf{R}_q[\mathbf{w}_q]_{\times}\mathbf{u}_i\\ \cdots \end{pmatrix}\\
    &= \begin{pmatrix} \cdots \\ [\mathbf{v}_i]_{\times}\mathbf{R}_r^T\mathbf{R}_q\mathbf{u}_i-[\mathbf{v}_i]_{\times}\mathbf{R}_r^T\mathbf{R}_q[\mathbf{u}_i]_{\times}\mathbf{w}_q\\ \cdots \end{pmatrix}\\
%
    \Bar{\mathbf{e}}_\mathbf{t} \lvert_{\mathbf{R}_r}
    &= \begin{pmatrix} \cdots \\ \lambda_i\mathbf{u}_i^T\mathbf{R}_q^T(\mathbf{t}_q-\mathbf{t})+\mathbf{u}_i^T\mathbf{R}_q^T\mathbf{R}_r(\mathbf{I}+[\mathbf{w}_r]_{\times})\mathbf{l} \\ \cdots \end{pmatrix} \\
    &= \begin{pmatrix} \cdots \\ \lambda_i\mathbf{u}_i^T\mathbf{R}_q^T(\mathbf{t}_q-\mathbf{t})+\mathbf{u}_i^T\mathbf{R}_q^T\mathbf{R}_r\mathbf{l} + \mathbf{u}_i^T\mathbf{R}_q^T\mathbf{R}_r[\mathbf{w}_r]_{\times}\mathbf{l} \\ \cdots \end{pmatrix} \\
    &= \begin{pmatrix} \cdots \\ \lambda_i\mathbf{u}_i^T\mathbf{R}_q^T(\mathbf{t}_q-\mathbf{t})+\mathbf{u}_i^T\mathbf{R}_q^T\mathbf{R}_r\mathbf{l} - \mathbf{u}_i^T\mathbf{R}_q^T\mathbf{R}_r[\mathbf{l}]_{\times}\mathbf{w}_r\\ \cdots \end{pmatrix} \\
    \Bar{\mathbf{e}}_\mathbf{t} \lvert_{\mathbf{R}_q}
    &= \begin{pmatrix} \cdots \\ \lambda_i\mathbf{u}_i^T(\mathbf{I}+[\mathbf{w}_r]_{\times})^T\mathbf{R}_q^T(\mathbf{t}_q-\mathbf{t})+\mathbf{u}_i^T(\mathbf{I}+[\mathbf{w}_r]_{\times})^T\mathbf{R}_q^T\mathbf{R}_r\mathbf{l} \\ \cdots \end{pmatrix} \\
    &= \begin{pmatrix} \cdots \\ \lambda_i\mathbf{u}_i^T(\mathbf{I}-[\mathbf{w}_r]_{\times})\mathbf{R}_q^T(\mathbf{t}_q-\mathbf{t})+\mathbf{u}_i^T(\mathbf{I}-[\mathbf{w}_r]_{\times})\mathbf{R}_q^T\mathbf{R}_r\mathbf{l} \\ \cdots \end{pmatrix} \\
    &= \begin{pmatrix} \cdots \\ \cdots - \lambda_i\mathbf{u}_i^T[\mathbf{w}_r]_{\times}\mathbf{R}_q^T(\mathbf{t}_q-\mathbf{t})+\cdots -\mathbf{u}_i^T[\mathbf{w}_r]_{\times}\mathbf{R}_q^T\mathbf{R}_r\mathbf{l} -\cdots \\ \cdots \end{pmatrix} \\
    &= \begin{pmatrix} \cdots \\ \cdots + \lambda_i\mathbf{u}_i^T[\mathbf{R}_q^T(\mathbf{t}_q-\mathbf{t})]_{\times}\mathbf{w}_r+\cdots +\mathbf{u}_i^T[\mathbf{R}_q^T\mathbf{R}_r\mathbf{l}]_{\times}\mathbf{w}_r -\cdots \\ \cdots \end{pmatrix} \\
    \end{aligned}
\end{equation}

In the linearized cost function $\mathbf{e}_{\mathbf{t}}|_{\mathbf{R}_q}$, constant terms not related to $\mathbf{x}$ are omitted.  

\section{Jacobians}
Finally, from the above linearized equations, we can have the Jacobian blocks: 
\begin{equation}
    \begin{aligned}
    \frac{\partial \mathbf{e}_\mathbf{R}}{\partial \mathbf{R}_r} &= \begin{pmatrix} \cdots \\ [\mathbf{v}_i]_{\times}[\mathbf{R}_r^T\mathbf{R}_q\mathbf{u}_i]_{\times} \\ \cdots \end{pmatrix}\\
    \frac{\partial \mathbf{e}_\mathbf{t}}{\partial \mathbf{R}_r} &= \begin{pmatrix} \cdots \\- \mathbf{u}_i^T\mathbf{R}_q^T\mathbf{R}_r[\mathbf{l}]_{\times}\\ \cdots \end{pmatrix}\\
    \frac{\partial \mathbf{e}_{\mathbf{R}_r}}{\partial \mathbf{R}_q} &= \begin{pmatrix} \cdots \\-[\mathbf{v}_i]_{\times}\mathbf{R}_r^T\mathbf{R}_q[\mathbf{u}_i]_{\times}\\ \cdots \end{pmatrix}\\
    \frac{\partial \mathbf{e}_\mathbf{t}}{\partial \mathbf{R}_q} &= \begin{pmatrix} \cdots \\\lambda_i\mathbf{u}_i^T[\mathbf{R}_q^T(\mathbf{t}_q-\mathbf{t})]_{\times} +\mathbf{u}_i^T[\mathbf{R}_q^T\mathbf{R}_r\mathbf{l}]_{\times} \\ \cdots \end{pmatrix}\\
    \end{aligned}
\end{equation}

As to Jacobian w.r.t. translation and scale, it is straight forward: 
\begin{equation}
    \begin{aligned}
    \frac{\partial \mathbf{e}_\mathbf{t}}{\partial \mathbf{t}_r} &=  \begin{pmatrix} \cdots \\-\lambda_i\mathbf{u}_i^T\mathbf{R}_q^T\\ \cdots \end{pmatrix}\\
    \frac{\partial \mathbf{e}_\mathbf{t}}{\partial \mathbf{t}_q} &=  \begin{pmatrix} \cdots \\\lambda_i\mathbf{u}_i^T\mathbf{R}_q^T\\ \cdots \end{pmatrix}\\
    \frac{\partial \mathbf{e}_\mathbf{s}}{\partial \mathbf{s}_q} &= \begin{pmatrix} \cdots \\ 2\mathbf{s}_i\\ \cdots \end{pmatrix}
    \end{aligned}
\end{equation}
The computed Jacobian blocks can then be filled into (\ref{eqn:jacobian}) and finally used to construct the complete Jacobian matrix used in the Levenberg Marquardt algorithm.